\documentclass[10pt,twocolumn,letterpaper]{article}

\usepackage[pagenumbers]{cvpr} 

\usepackage{graphicx}
\usepackage{amsmath}
\usepackage{amssymb}
\usepackage{booktabs}
\usepackage{algorithm}
\usepackage{algorithmic}
\usepackage{float}
\usepackage{bm}
\usepackage{bbm}
\usepackage{placeins}
\usepackage{stfloats}
\usepackage{multirow}
\usepackage{makecell}
\usepackage{utfsym}
\usepackage{diagbox}
\definecolor{cvprblue}{rgb}{0.21,0.49,0.74}
\usepackage[pagebackref,breaklinks,colorlinks,allcolors=cvprblue]{hyperref}

\def\paperID{2303} 
\def\confName{CVPR}
\def\confYear{2025}
\newcommand{\method}{TASE}
\newcommand{\methodspace}{TASE }
\title{Dataset Awareness is not Enough: Implementing Sample-level Tail Encouragement in Long-tailed Self-supervised Learning}

\author{Haowen Xiao$^{1}$, Guanghui Liu$^{3}$, Xinyi Gao$^{1}$, Yang Li$^{1}$, Fengmao Lv$^{2}$, Jielei Chu$^{2}$\\
\textsuperscript{1}{School of Information Science and Technology, Southwest Jiaotong University}\\
\textsuperscript{2}{School of Computing and Artificial Intelligence, Southwest Jiaotong University}\\
\textsuperscript{3}{College Of Computer Science And Technology, Harbin Engineering University}\\
{\tt\small Xhaowen@my.swjtu.edu.cn, fengmaolv@126.com, jieleichu@swjtu.edu.cn}}

\begin{document}
\maketitle

\begin{abstract}
Self-supervised learning (SSL) has shown remarkable data representation capabilities across a wide range of datasets. However, when applied to real-world datasets with long-tailed distributions, performance on multiple downstream tasks degrades significantly. Recently, the community has begun to focus more on self-supervised long-tailed learning. Some works attempt to transfer temperature mechanisms to self-supervised learning or use category-space uniformity constraints to balance the representation of different categories in the embedding space to fight against long-tail distributions. However, most of these approaches focus on the joint optimization of all samples in the dataset or on constraining the category distribution, with little attention given to whether each individual sample is optimally guided during training.
To address this issue, we propose Temperature Auxiliary Sample-level Encouragement (TASE). We introduce pseudo-labels into self-supervised long-tailed learning, utilizing pseudo-label information to drive a dynamic temperature and re-weighting strategy. Specifically, We assign an optimal temperature parameter to each sample. Additionally, we analyze the lack of quantity awareness in the temperature parameter and use re-weighting to compensate for this deficiency, thereby achieving optimal training patterns at the sample level.
Comprehensive experimental results on six benchmarks across three datasets demonstrate that our method achieves outstanding performance in improving long-tail recognition, while also exhibiting high robustness.
\end{abstract}    

\section{Introduction}
\label{sec:intro}

\begin{figure}[!ht]
  \centering
   \includegraphics[width=0.8\linewidth]{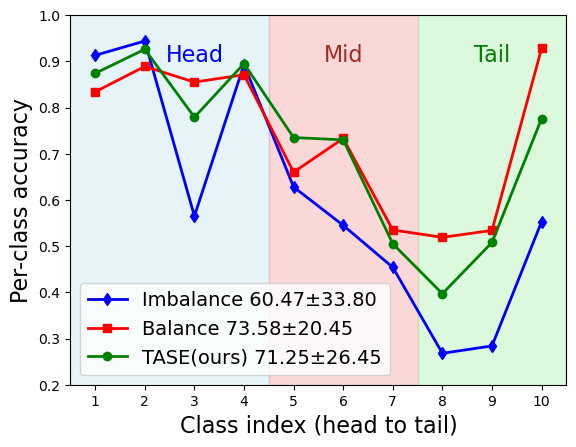}
   \caption{Comparison between SimCLR \cite{SimCLR} baseline trained on balanced CIFAR10\cite{cifar} and imbalanced CIFAR10-LT \cite{cifar_LT} and our \methodspace   trained on CIFAR10-LT. Three Splits (Head, Mid and Tail) are shown as different colours (blue, red and green) respectively \cite{liu2019large}. We also report the average accuracy and range of variation over ten classes.}
   \label{fig:motivation}
\end{figure}

Self-supervised learning (SSL) \cite{MOCO,MAE}, which aims to learn robust and generic feature representations from unlabeled data, has been consistently explored in recent years \cite{SimSiam,SwAV,Barlow_Twins}. Contrastive learning, a leading SSL paradigm, seeks to learn discriminative representations by encouraging the embedding similarity of augmented view pairs, and has achieved remarkable success across several domains \cite{ImGCL,BYOL}. Nevertheless, most contrastive learning methods are conducted on curated datasets like ImageNet \cite{imagenet}. In contrast, real-world data distributions are mainly characterized by long-tailed distributions \cite{distribution}, which could cause a significant degradation in performance, as illustrated in \cref{fig:motivation}. On the relatively simple CIFAR10-LT dataset, the baseline model demonstrates a notable performance collapse. For certain tail classes, the accuracy even falls below 30\%, indicating that SSL still suffers from the long-tailed problem. The difficulty of detecting or modeling such implicit imbalance is a significant obstacle to migrating SSL to large-scale, uncurated, real-world datasets.

In recent years, for the supervised paradigm, the long-tailed problem has been extensively studied from multiple perspectives, deriving strategies such as re-weighting \cite{BCL_SCL}, re-sampling, and ensemble learning, which have achieved significant success. In terms of re-weighting, a common approach is to increase the weight proportion of tail-class samples, in order to occupy a larger gradient and encourage the model to focus on tail-class samples. Turning to self-supervised learning, particularly contrastive learning, some works have proposed using temperature assignment to counteract the effects of long-tailed distributions \cite{Behaviour_of_Contrastive_Loss_tau, TS}. Specifically, models can be made to focus on different semantics by adjusting the temperature parameters of the contrastive objective. A lower temperature encourages sample features to push similar samples away as much as possible, independently forming semantic clusters. This is a pattern emphasizes category discrimination, while a higher temperature allows greater tolerance for similar samples, focusing more on semantic difference learning and instance discrimination. Such learning pattern difference is a potential way to combat long-tailed distributions.

Then our perspective shifts back to self-supervised long-tailed learning, some research has sought to adapt the aforementioned mechanisms to this domain. TS \cite{TS} attempts to employ a global temperature with cosine variation to modify the model's learning pattern between instance discrimination and category discrimination. GH \cite{GH} emphasizes the necessity for spatial homogeneity in imbalanced learning scenarios and proposes a plug-and-play objective for harmonizing the embedding space of each category. Nevertheless, none of these efforts achieve optimal training for all samples, and intuitively, it is more efficient to ensure that each sample is appropriately trained than to constrain the global.

Motivated by the above considerations, we notice pseudo-labels assignment in semi-supervised learning, and then use pseudo-labels of each sample to achieve sample-level encouragement. 
Nevertheless, due to the long-tailed distribution, the tail classes with few samples are more likely to yield severe misassignments. In order to avoid serious misleading of the tail classes, several pseudo-labels assignment methods from semi-supervised learning, which use pseudo-labels as targets for consistency regularization \cite{sohn2020fixmatch, zhang2021flexmatch}, cannot be directly transferred to the long-tailed scenario. Therefore, we make the following attempts and propose \method. 
Specifically, we assign appropriate temperature based on the category to which the pseudo-labels of samples belong. This allows the head class to learn semantic differences from other similar samples, and while the tail class is separated from other sample clusters, thereby generating individual categories. Nevertheless, during the training process, the head and tail classes are assigned the maximum and minimum temperatures respectively, but the range of temperature parameters is often treated as a fixed parameter, which results in the strength of tail encouragement having the same intensity for datasets with different imbalance ratios. We call this as quantity awareness flaw of temperature strategy.
To address this flaw, we propose re-weighting strategy to provide explicit quantity awareness. In particular, we add a weight to each sample that is inversely proportional to the number of samples in the category to which it belongs.

Compared to existing work, our method has the following advantages: 1) Compared to previous work on TS \cite{TS} and GH \cite{GH}, we are able to modify the training process at the sample-level, providing the most appropriate encouragement to each sample to counteract the long-tailed distribution. 2) Compared to the approach in semi-supervised learning, where pseudo-labels are directly used as targets for consistency regularization, our method does not require pseudo-labels for direct training, and it is better at resisting the misleading effects of incorrect label assignments.

\noindent
We summary contributions of this paper as below:  
\begin{itemize}
    \item We note the drawback of existing work on long-tail self-supervised learning that it is difficult to perform tail encouragement for the sample level, and propose to introduce the idea of pseudo assignment into long-tail self-supervised learning to perform sample-specific learning for each sample.
    \item We propose \methodspace as a preliminary attempt to implement sample-level control in long-tail self-supervised learning, assigning appropriate temperatures driven by pseudo assignments for different samples during training to achieve optimal training patterns.
    \item We implement comprehensive experiment across six benchmarks on three commonly used datasets, and the experimental results demonstrate our model achieves superior performance.
\end{itemize}

\section{Related Work}
\label{sec:related_work}

\noindent
{\bf Supervised Long-tailed Recognition.} Classical supervised long-tail recognition can be categorized into three directions: category-level re-balancing, information augmentation and module improvement \cite{zhang2023deep}. Related to our work, exploration of category-level re-balancing mainly focuses on re-sampling, re-weighting and logit adjustment \cite{decoupling,BCL_SCL,VS,adaptive}. Decoupling \cite{decoupling} explores the impact of different sampling strategies, and finds that square-root sampling \cite{squareroot} and progressively-balanced sampling are better suited for long-tailed datasets. LOCE \cite{Loce} proposes a memory-agumented feature sampling to dynamically monitor the prediction score of each classes to drive re-sampling rate distribution. Oriented towards the supervised contrastive learning paradigm, BCL \cite{BCL_SCL} develops a novel loss for balanced training, which considers the averaging of negative class gradients and the complement of mini-batch tail classes. Logit adjustment \cite{logit-adjustment} utilises training label frequencies to adjust the prediction logit. VS loss \cite{VS} conducts a comprehensive analysis on the impact of additive and multiplicative logit-adjusted losses, and further formulates a novel strategy that combines the best of both worlds.

Explorations on information augmentation seek to introduce additional information into training process, mainly containing transfer learning and data augmentation \cite{ssp, MiSLAS, han2005borderline, ssd, deitlt, gou2021knowledge}. SSP \cite{ssp} systematically analyzes the benefits of combining semi-supervised and self-supervised manners to standard learning for balanced feature space learning. SSD \cite{ssd} develops a soft labels generator driven by self-supervised distillation, producing robust auxiliary information to transfer knowledge from long-tailed distribution. DeiT-LT \cite{deitlt} aims to tackle the problems of high efficiency training from scratch, introducing an effective distillation construct from a flat CNN teacher via CLS token and distillation DIST token to support a head expert and a tail expert, respectively. Remix \cite{remix} benefits from re-balanced data mixup to boost long-tail learning. SBCL \cite{Sub_CL} employs subclass-balancing adaptive clustering and bi-granularity contrastive loss to facilitate the implicit mining of sub-classes and optimise the balance of sub-classes.

Module improvement can be divided into classifier design and ensemble learning \cite{paco, MiSLAS, kang2020exploring, RIDE, SADE}. Paco \cite{paco} introduces a set of parametric class-wise learnable centers to  adaptively enhance the optimisation of latent space. RIDE \cite{RIDE} leverages a strategy of multiple experts driven by KL-divergence based loss to benefit hard example learning. SADE \cite{SADE} explores a multi-expert scheme to handle test-agnostic long-tailed recognition, innovating diversity-promoting expertise-guided losses to adaptively aggregate experts for handling unknown distribution.

\noindent
{\bf Self-supervised Long-tailed Recognition} expects to automatically correct potential category bias without ground-truth labels and data distributions. There are several recent explorations for this purpose. SDCLR \cite{SDCLR} constructs a pruned dynamic self-competitor for mining forgettable samples, which are then implicitly emphasized in the contrastive process. DnC \cite{DnC} alternates between base contrastive learning and multi-experts hard negative mining, aggregating features from a distillation step to mitigate the collapse of the rare classes. TS \cite{TS} leverages an oscillating temperature factor with a cosine schedule to dynamically learn both group-wise and instance-specific features. FASSL \cite{FASSL} presents a frequency-aware prototype learning to identify prototypes guidance reflecting latent distribution from unlabeled data, benefiting re-balanced learning. From a data perspective, BCL-I \cite{BCL_SSL} exploits the memorization effect of DNNs to automatically drive an augmentation of each instance, thereby aiding the learning of discriminative features. GH \cite{GH} develops a novel contrasting loss from the perspective of geometric harmonisation, dynamically approaching the class-level uniformity of the latent space.

\noindent
{\bf Pseudo-labels Assignment.} DeepCluster \cite{caron2018deep} leverages the cluster assignments as pseudo-labels, driving an end-to-end unsupervised learning with the soft labels classification pretext task. Subsequently, a large number of methods have been developed to address degenerate solutions, i.e., to prevent collapse. Unlike end-to-end approaches, SCAN \cite{scan} employs a method that decouples representation learning from clustering, ensuring that the clustering process does not depend on low-level features. SeLa \cite{asano2019self}, on the other hand, focuses on maximizing the information between labels and input data indices. SwAV \cite{caron2020unsupervised} develops a method that maps images to a set of trainable prototype vectors, which simultaneously avoids the computational cost of pairwise comparisons and improves the model's scalability and training efficiency.

\section{Method}
\label{sec:method}
\subsection{Preliminary}
\noindent
For a given dataset $\mathcal{D}$, every instance is denoted by $(\bm{x},\bm{y})\in \mathcal{D}$,where $\bm{y}$ is corresponding class label of input $\bm x$. Contrastive learning expects to close similarity of positive pairs $(\bm{x_i}, \bm{x_{i'}})$ produced by a random augmentation and repel similarity of negative pairs $(\bm{x_i}, \bm{x_{j}})$. The contrastive loss for instance $i$ is defined by:
\begin{equation}
\label{eq:loss}
    \ell_{i} = -\log \frac{S(\bm v_i, \bm v_{i'},\tau)}{S(\bm v_i, \bm v_{i'},\tau)+\sum_{v_j\in\mathcal{B}^-_i} S(\bm v_i, \bm v_j,\tau)}~
\end{equation}
\begin{equation}
\label{eq:S}
    S(\bm v_i, \bm v_j,\tau) = \exp(sim(\bm v_i, \bm v_j)/\tau) ~
\end{equation}
Where $sim(\cdot)$ denotes the cosine similarity, $B^-_i$ denotes a set of negative samples of $x_i$ and $v_i=g(f(x_i))$ is embedding of $x_i$. $f(\cdot)$ and $g(\cdot)$ are encoder and projection head network, respectively.

\subsection{Temperature Property}
\label{subsec:temperature}
In recent years, research has been conducted to examine the mechanisms of different temperature parameters in greater details \cite{robinson2021can,Behaviour_of_Contrastive_Loss_tau}. Several evidence from both quantitative and qualitative experiments has demonstrated that temperature parameters exert a profound influence on the uniformity and tolerance of training process. Previous work \cite{Behaviour_of_Contrastive_Loss_tau} calls these two opposing properties as the uniformity-tolerance dilemma, corresponding to instance-discrimination and category-discrimination, respectively. To be specific, for \cref{eq:loss} and \cref{eq:S}, given a small $\tau$, the penalty will be dominated by hard negative samples, and therefore more attention will be paid to push the more similar samples farther away, \ie, the tolerance of this feature point is small. Accordingly, when the tolerance of all the points is small, the latent space will be paved over as much as possible, presenting a high uniformity feature. In contrast, a large $\tau$ exhibits high tolerance and low uniformity; therefore, feature points will try to occupy as little space as possible. 

We give a gradient analysis to illustrate this. The gradient for the similarity between $v_i$ and a random negative sample $v_k$ is denoted as \cref{eq:gradient}, and the ratio of its gradient to the sum of gradients with respect to all negative samples is given by \cref{eq:ratio}. When $sim(\boldsymbol{v}_i,\boldsymbol{v}_k)$ tends to 1, \ie, $v_k$ is a hard negative example, with the value of $\tau$ linearly decrease, $r_k$ will increase significantly, implying that the hard negative samples gradually dominates gradient direction.

\begin{equation}
\label{eq:gradient}
    \frac{\partial\mathcal{L}(x_i)}{\partial sim(\boldsymbol{v}_i,\boldsymbol{v}_k)}=\frac1\tau \frac{S(\boldsymbol{v}_i,\boldsymbol{v}_{k},\tau)}{S(\boldsymbol{v}_i,\boldsymbol{v}_{i'},\tau)+\sum_{\boldsymbol{v}_j\in\mathcal{B}_i^-}S(\boldsymbol{v}_i,\boldsymbol{v}_j,\tau)}
\end{equation}

\begin{equation}
\label{eq:ratio}
    r_k=\frac{\exp(sim(\boldsymbol{v}_i,\boldsymbol{v}_k))/\tau)}{\sum_{\boldsymbol{v}_j\in\mathcal{B}_i^-}\exp(sim(\boldsymbol{v}_i,\boldsymbol{v}_j)/\tau)}
\end{equation}

\subsection{Structural Separability and Category Uniformity}
\label{structure}
In this section, we provide an analysis about the expectation of the semantic spatial structure and how the dynamic temperature strategy could help reach this ideal structure.

\begin{figure}[!ht]
  \centering
   \includegraphics[width=0.8\linewidth]{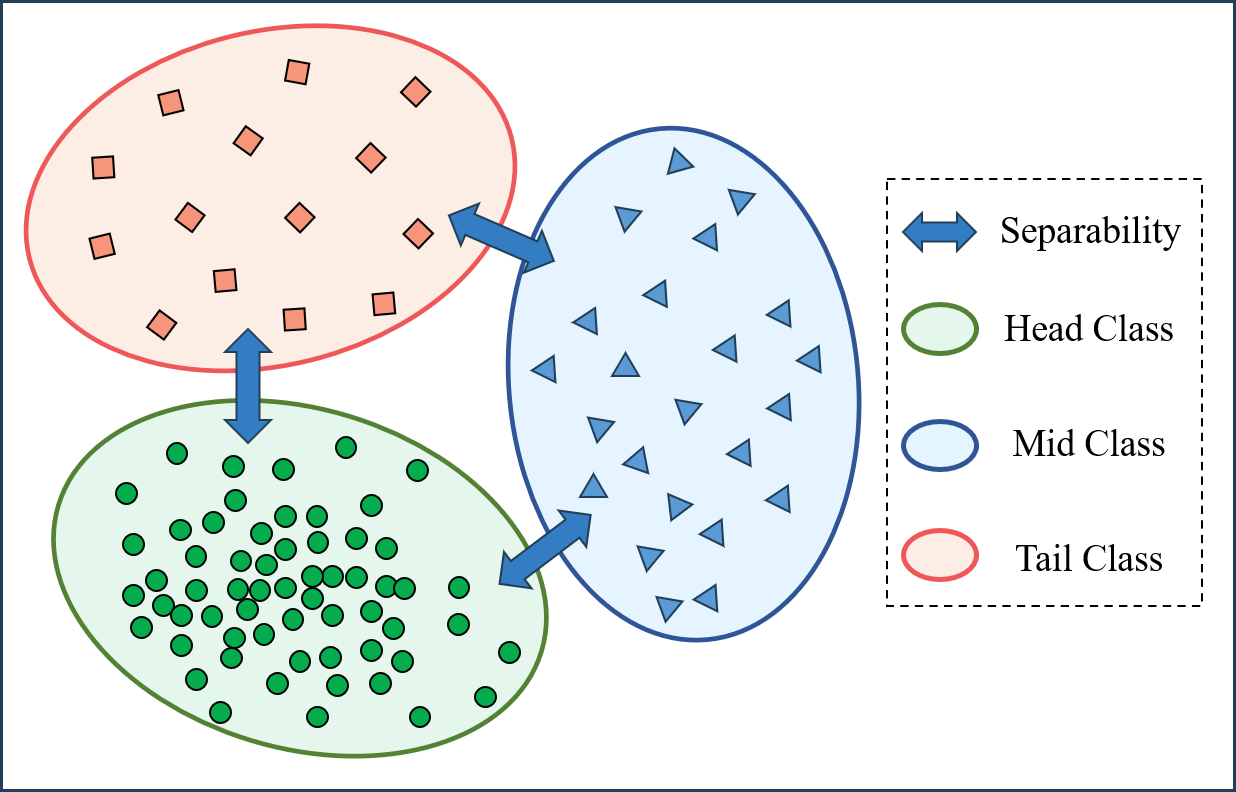}
   \caption{Illustration of the structure of the desired embedding space. We use different icons in different colors to represent the samples from different categories and indicate the size of the space occupied by each category. We expect the clusters to be at a certain distance from each other and to occupy a similar amount of space.}
   \label{fig:balance}
\end{figure}

In order to obtain robust downstream task performance, higher requirements are demanded for the semantic space structure of the data representation, which we describe as structural separability and category uniformity. As shown in \cref{fig:balance}, structural separability expects that samples from different categories form category semantic clusters and maintain strong linear separability. At the same time, in order to focus on the tail categories, category uniformity expects that each category occupying embedding space should be similar, irrespective of the number.

From the perspective of temperature setting, a lower temperature can make the sample pay more attention to strong negative examples and pursue better separation. On the other hand, stronger uniformity can make the latent space occupied by each category larger, which is more suitable for categories with fewer samples. Conversely, we hope to set a higher temperature for categories with more samples, encouraging the model to learn the inherent semantic differences between samples and maintain high generalization across a variety of transfer learning tasks. To this end, we utilize a temporary feature clustering module to assign dynamic pseudo-labels as auxiliary information to guide the assignment of temperature. If the sample belongs to the head class, a higher temperature is set, and if it belongs to the tail class, a lower temperature is assigned.
There is no restriction on the clustering method. We use a simple K-means to verify our idea.

\subsection{Explicit Quantity Awareness}
\label{reweight}
In \cref{structure}, we describe how we use dynamic temperature tuning strategy to optimize spatial structure. However, we find this to be a flaw in practice, as the general temperature settings are treated as fixed hyperparameters, and pseudo-labels cannot provide information on imbalance ratios. In other words, this flaw gives rise to the fact that the same temperatures will always be assigned to the head class samples on different datasets with different imbalanced ratio. Therefore, we require an adjunct that can take advantage of the information on the distribution of pseudo-labels to jointly serve as a hybrid objective.

Therefore, we introduce a classic supervised re-weighting strategy to assist category uniformization. For each sample, as negative examples, a weight is added inversely proportional to the number of samples in the category to which they belong. And to prevent overcorrection we use the inverse of the square root of the sample size as the weights.

\subsection{Training via \method}
\label{loss}
In this section, we introduce our design of loss function feature and clustering component.
\begin{equation}
\label{eq:cross_loss}
     \ell_{i} = -\log \frac{S(\bm v_i, \bm v_{i'},\tau_i)}{S(\bm v_i, \bm v_{i'},\tau_i)+\sum_{\boldsymbol{v}_j\in\mathcal{B}^-_i}\frac {1}{\sqrt{|D_j|}} S(\bm v_i, \bm v_j,\tau_i)}~
\end{equation}
\begin{equation}
\label{eq:cross_loss}
     \mathcal{L}_{TASE} = \frac{1}{|D|}\sum_{\boldsymbol{v}_i\in D}{\ell_i}
\end{equation}
 First, we give our hybrid cross-paradigm loss as \cref{eq:cross_loss}, where $\tau_i$ denotes a dynamic temperature assigned by temporary labels, and $|D_j|$ denotes the number of the samples that is allocated to the same class as $v_j$.

Our training is divided into two stages: warming for $B$ epochs and cluster balance training for the rest epochs. In the warming phase, we utilize original SimCLR to seek a good cluster initialization. After $B$ epochs, we use a progressive temperature factor until $S$ epochs.

The pseudo-code is summarized as \cref{algorithm:method}
\begin{algorithm}[!htb] 
  \caption{Pseudo-code of our proposed \method.} \label{algorithm:method}
    {\bf Input:} dataset $\mathcal{D}$, number of epochs $E$, number of warming epochs $B$, cluster frequency $F$, encoder network $f$ \\
    {\bf Output:} pretrained model parameter $\theta_{E}$\\
    {\bf Initialize:} model parameter $\theta_{0}$
\begin{algorithmic}[1]
    \STATE Warm up model $\theta_0$ for $B$ epochs via SimCLR.
    \STATE Initialize the pseudo-label allocation $M$. 
    \FOR{epoch $e=B,B+1,\ldots,E$}
    \STATE Compute the distribution of temporary classes assignment $P$ by $M$.
    \FOR{mini-batch $n=1,2,\ldots,N$}
    \STATE Obtain the temperature $\mathbf{t}$ for each sample by $P$ .
    \STATE Compute the proposed $\mathcal{L}_{\mathrm{\method}}$.
    \STATE Uptate model $\theta$ by $\mathcal{L}_{\mathrm{\method}}$.
    \ENDFOR
    \IF{epoch $e$ \% $F$ == 0}
    \STATE Update  pseudo-label allocation $M$.
    \ENDIF
    \ENDFOR

\end{algorithmic}
\end{algorithm}

\subsection{More Discussion}
\noindent
\textbf{Computational Complexity and Storage Consumption.} Our method maintains a temporary list of pseudo-labels with length equal to the number of samples and performs a cross-batch clustering operation once in a certain number of rounds. Our method requires only a small increase in storage consumption and computational overheads.

\noindent
\textbf{Expandability} Our \methodspace can be combined with other contrastive learning frameworks \cite{BYOL,MOCO} or non-contrastive self-supervised frameworks \cite{MAE}. Correcting potential imbalances by using only temporary image representations results in high availability and low overhead. The feasibility and extensibility of \methodspace have been validated by combining it with SimCLR, and we will make combining it with other self-supervised learning frameworks in the future.

\section{Experiment}
\label{sec:experiment}
\subsection{Datasets}
\label{subsec:datasets}
\noindent
{\bf CIFAR10-LT and CIFAR100-LT \cite{cifar_LT}} are long-tailed versions of the CIFAR10 and CIFAR100 datasets, respectively. The original CIFAR10 \cite{cifar} dataset comprises of 32×32 images from 10 classes, with 50k images allocated for training and 10k images for testing. The imbalance factor is defined as the number of the head classes divided by the number of tail classes. Following SDCLR \cite{SDCLR} and TS \cite{TS}, we set the imbalance factor to 100. Each class contains between 4,500 and 45 images according to the Pareto distribution. In total, there are 11,165 images in the training and test sets, with 10,000 images remaining.

Similarly, the CIFAR100 dataset has 100 classes of images with 50k images for training and 10k for testing. The number of images per class for training in CIFAR100-LT varies from 450 to 4, with a total of 9,754 images. We average all of our results over three runs across five splits to reduce randomness to the greatest extent possible.
\\{\bf ImageNet100-LT} is a subset of ImageNet100 \cite{CMC_imagenet100} dataset, consisting of 12.21k images from 100 classes for training and 5k images for testing. The number of images of each class in the training set range from 1280 to 5 according to the Pareto distribution.

\subsection{Experimental Setup}
\label{subsec:setup}
\noindent
{\bf Evaluation.} TS \cite{TS} provides a comprehensive validation that employs K nearest neighbours \cite{knn} (KNN) to verify the latent space structure by directly evaluating the learned local representations, and two linear probing (LP) methods to validate the generalization and linear separability \cite{SDCLR} of learned embedding, which we call the balanced minimum sampling linear probing (MS LP) and the long-tailed linear probing (LT LP). We follow their validation methods, and in addition, we provide two other more widely used validation methods, the 1\% sampling linear probing (1\%S LP) and the full set linear probing (Full LP), all of which we will explain in detail below.

KNN \cite{knn} can directly evaluate the learned spatial structure. We compute L2-normalized distances between training samples and testing samples. The class of each testing image is then predicted as the majority class among the top-K closest training samples. Two series of accuracy have been established for KNN with \(K=1\) and \(K=10\) marked as KNN@1 and KNN@10, respectively.

Linear probing \cite{linearprobe} trains a linear classifier layer on the top of pre-trained feature extractor by using specified data and then evaluates accuracy on the testing set. By default, we use full set as fine-tune dataset and validate it on the testing set called Full LP.
Considering practical scenarios, we also consider two balanced few shot setups \cite{fewshot}: minimum sampling linear probing (MS LP) and 1\% sample linear probing (1\%S LP), both use 1\% of the full dataset and have an equal number of samples between each class. The difference is that the former samples the data in the pre-train dataset as a fine-tuning dataset with better uniformity, while the latter randomly samples 1\% of all the dataset, which is more closely aligned with the actual scenario. Additionally, we also conduct evaluation driven by imbalanced fine-tuning dataset called long-tailed linear probing (LT LP), using original imbalanced training set.

\begin{table*}[ht]
\centering
\normalsize
\renewcommand{\arraystretch}{0.95}
\tabcolsep=0.2cm
\begin{tabular}{c|cccccc}
\textbf{Method} & KNN@1  & KNN@10 & MS LP & 1\%S LP & LT LP  & Full LP\\
\midrule
\multicolumn{7}{c}{\textbf{ResNet18}}\\
\midrule
SimCLR & 59.84 & 60.19 & 68.29 & 62.98 & 61.86 & 75.37 \\
SDCLR & 61.92 & 60.63 & 71.74 & 71.74  & 65.38 & 80.49\\
TS & 63.09 & 62.91 & 71.86 & 71.29  & 65.03 & 79.08\\
FASSL & \textbf{---} & \textbf{---} & \textbf{---} & 68.54  & \textbf{---} & \textbf{80.69}\\
\method & \textbf{64.06} & \textbf{63.70} & \textbf{72.78} & \textbf{72.60}  & \textbf{65.46} & 80.35\\
\midrule
\multicolumn{7}{c}{\textbf{ResNet50}}\\
\midrule
SimCLR &62.51&61.49&72.83&72.17&65.78&83.75 \\
SDCLR & 62.59 & 61.27 & 73.30 & 72.73 & 67.07 & \textbf{85.59}\\
TS & 63.11&61.94&73.54&73.28&66.80&84.76\\
\method& \textbf{64.17}&\textbf{62.83}&\textbf{74.16}&\textbf{74.27}&\textbf{68.15}&85.46\\
\bottomrule
\end{tabular}
\caption{Performance comparison on CIFAR10-LT dataset. We utilizes Resnet18 and Resnet50 as backbone to reproduce some recent long-tailed self-supervised methods, and we bold the best performance.}
\label{tab:cifar10}
\end{table*}

\noindent
{\bf Implementation Details.} Following previous works \cite{SDCLR, TS}, in the case of smaller-scale datasets CIFAR10-LT and CIFAR100-LT, the ResNet-18 architecture is employed as the backbone for training over 2000 epochs, while for the ImageNet100-LT dataset, the ResNet-50 is utilised for training over 800 epochs. Additionally, we train our model with SGD optimizer, batchsize of 512 for CIFAR10-LT and CIFAR100-LT, 256 for ImageNet100-LT, momentum of 0.9, feature dimension of 128 and weight decay of \(1e^{-4}\). We adopt a learning rate followed by cosine annealing schedule with 10 epochs linear warm-up and set learning rate as 0.5 for CIFAR and 2 for ImageNet.

As for our unique hyperparameters, The value of clusters \(K\) is set as equal to the number of classes contained in the dataset, \ie, \(K\)=10 for CIFAR10-LT and 100 for ImageNet100-LT and CIFAR100-LT. The value of B is equivalent to the starting epoch of clustering, i.e. B=50 for CIFAR10-LT and ImageNet100-LT, 300 for CIFAR100-LT. In all three datasets, we assign the frequent of clustering F to 10. Epochs of warming S=500 for CIFAR10-LT, 1000 for CIFAR100-LT, and 200 for ImageNet100-LT. We will have an in-depth analysis of parameters sensitivity and detailed experimental results in \cref{subsec:sensitivity}.

To gain a more detailed insight of the quantitative impact of our method for calibrating domination of majority classes, we divide all classes of each dataset into three groups from highest to lowest sample size: Head, Mid and Tail. For CIFAR10-LT and CIFAR100-LT datasets, we split the three groups evenly, 4:3:3 for CIFAR10-LT and 34:33:33 for CIFAR100-LT. As for ImageNet100-LT dataset, following \cite{liu2019large}, defining head classes as classes with over 100 samples, and tail as classes under 20 samples, classes have the
number of samples from 20 to 100 are assigned as Mid classes.

\subsection{Experimental Results}
\label{subsec:results}
In this section, we summarize the comprehensive experimental results by comparing our \methodspace with recent methods across a range of benchmarks, and offer insights into the observed outcomes.

In the CIFAR10-LT and CIFAR100-LT datasets, we conduct comparison experiments across all six benchmarks previously mentioned. As shown in \cref{tab:cifar10} and \cref{tab:cifar100}, for comparison of latent space structure with the ResNet18 backbone model, \ie, KNN@1 and KNN@10, we achieve a significant consistent improvement compared to previous methods, which demonstrates that our approach efficiently optimises local embedding representations, which are based on the learned metric. In terms of practical considerations, we adopt three settings of linear probing experiment: 1\%S LP, MS LP and LT LP. We find that the average accuracy is positively correlated with the number of fine-tuned samples. For the MS LP and 1\%S LP benchmarks, the proposed method continues to demonstrate satisfactory performance. Differences in data intersections for training and fine-tuning yielded by the two settings exerts the most significant influence on SimCLR, while has a limit impact on other methods. At setting of LT LP, we utilize original long-tailed pre-training set for fine-tune. Compared to MS LP and 1\%S LP, LT LP has a greater uniformity and more fine-tuning instances. Despite the reduction in the lead, our \methodspace still achieves a competitive result. But for the settings of Full LP, our \methodspace does not perform as expected, especially in the CIFAR100-LT dataset.

\begin{table*}[!ht]
\centering
\normalsize
\renewcommand{\arraystretch}{0.95}
\tabcolsep=0.25cm
\begin{tabular}{c|cccccc}
\textbf{Method} & KNN@1  & KNN@10 & MS LP & 1\%S LP & LT LP  & Full LP\\
\midrule
\multicolumn{7}{c}{\textbf{ResNet18}}\\
\midrule
SimCLR & 28.81 & 28.12 & 25.70 & 22.51 & 31.2 & 49.76 \\
SDCLR& 30.19 & 29.59 & 26.63 & 27.32 & 34.38 & 55.00\\
FASSL & \textbf{---} & \textbf{---} & \textbf{---} & 27.11  & \textbf{---} & 55.27\\
TS & 31.06 & 30.06 & 28.89 & 28.32  & 33.28 & 51.64\\
GH & 29.34 & 28.58 & 27.73 & 27.60 & 34.13 & 53.18\\
BCL-I & 30.52 & 30.07 & 28.62 & 28.49 & \textbf{34.99} & \textbf{55.79} \\
\method & \textbf{31.27} & \textbf{30.81} & \textbf{29.02} & \textbf{28.69} & 33.96 & 52.21\\
\midrule
\multicolumn{7}{c}{\textbf{ResNet50}}\\
\midrule
SDCLR & 30.17 & 29.57 & 26.15 & 25.71 & 35.67 & 59.43 \\
TS & 31.50& 30.60&	28.46&	28.74&	36.00& 	58.81 \\
BCL-I & 30.38 & 29.68 & 28.68 & 29.07 & 36.62 & 59.30 \\
\method & \textbf{32.14} & \textbf{31.34} & \textbf{29.19} & \textbf{29.17} & \textbf{36.73} & \textbf{59.57}\\
\bottomrule
\end{tabular}
\caption{Performance comparison on CIFAR100-LT dataset. We utilizes Resnet18 and Resnet50 as backbone to reproduce some recent long-tailed self-supervised methods, and we bold the best performance.}
\label{tab:cifar100}
\end{table*}

\begin{table}[ht]
\centering
\small
\tabcolsep=0.2cm
\begin{tabular}{c|ccccc}  
\textbf{Method} & K@1 & K@10 & 1\%S LP & LT LP  & Full LP\\
\midrule
SimCLR & 37.10 & 38.00 & 42.16 & 44.82 & 65.46\\
SDCLR & 36.06 & 37.36 & 42.38 & 46.40 & 66.48\\
TS & 38.36 & 38.86 & 45.18 & 47.26 & 67.82\\
\method & \textbf{39.54} & \textbf{40.26} & \textbf{45.48} & \textbf{47.82} & \textbf{68.30}\\
\bottomrule
\end{tabular}
\vspace{.25em}
\caption{Performance comparison on ImageNet100-LT dataset. The best performance is highlighted in bold.}
\label{table:imagenet100}
\vspace{-1em}
\end{table}

In order to gain a deeper understanding of whether this performance discrepancy persists under different alternative models, a series of analogous experiments were conducted around ResNet50. We summarise the experimental results in the ResNet50 section of \cref{tab:cifar10} and \cref{tab:cifar100}. The experimental results indicate that our method does not result in performance collapse when utilising ResNet50, and maintains an excellent performance like KNN series in settings of ResNet18 backbone. It is worth noting that both temperature adjustment methods (\ie, TS and \method) yield the similar performance collapse, and this crash appears to be exclusive to the CIFAR100-LT dataset, which features a greater number of categories, and was more pronounced when utilising more fine-tuned data.

\begin{table}[ht]
\centering
\small
\renewcommand{\arraystretch}{1.1}
\tabcolsep=0.13cm
\begin{tabular}{c|cccccc} 
$\bm{K}$ &  K@1  & K@10 & MS LP & 1\%S LP & LT LP  & Full LP\\
\midrule
\textbf{70}  & \underline{30.46}&	\textbf{29.46}&	27.37&	\textbf{29.41}&	33.23&	\textbf{52.54}\\
\textbf{100} & \textbf{30.76}&	29.31&	\textbf{28.54}&	\underline{28.45}&	\underline{32.80}&	\underline{51.82}\\
\textbf{130} & 30.66&	\underline{29.19}&	27.47&	28.51&	\textbf{33.53}&	52.40\\
\textbf{200} & 30.69&	29.36&	\underline{27.31}&	28.46&	33.27&	51.84\\
\midrule
$R$ & 0.3 & 0.27 & 1.23 & 0.96 & 0.73 & 0.72\\
\end{tabular}
\vspace{.25em}
\caption{Effect of $\bm{K}$ on CIFAR100-LT dataset with ResNet18. $\bm{K}$ denotes the number of clusters. The best performance is highlighted in bold, while the worst performance is underlined. Additionally, the range of fluctuations has been indicated with $R$.}
\label{table:k_sensitivity}
\vspace{-1em}
\end{table}

Further, we also summary performance comparison on ImageNet100-LT dataset with ResNet50, as shown in \cref{table:imagenet100}.

\begin{table}[h]
\centering
\small
\renewcommand{\arraystretch}{1.1}
\tabcolsep=0.16cm
\begin{tabular}{c|cccccc} 
$\bm{B}$ &  K@1  & K@10 & MS LP & 1\%S LP & LT LP  & Full LP\\
\midrule
\textbf{0}   & \underline{62.77}&	\underline{61.50}&	71.61&	\textbf{71.29}&	63.58&	79.72\\
\textbf{50}  & 63.08& 	\textbf{62.30}&	\textbf{71.78}& 	70.84&	\textbf{64.50}& 	79.58\\
\textbf{100} & 63.01&	61.67&	\underline{71.10}&	70.55&	\underline{63.16}&	\underline{79.29}\\
\textbf{300} & \textbf{63.36}&	62.04&	71.40&	\underline{70.19}&	63.46&	\textbf{79.76}\\
\textbf{500} & 63.14&	61.82&	71.36&	70.77&	64.43&	79.35\\
\midrule
$R$ & 0.59 & 0.8 & 0.68 & 1.1 & 1.34 & 0.47\\
\end{tabular}
\vspace{.25em}
\caption{Effect of $\bm{B}$ on CIFAR10-LT dataset with ResNet18. $\bm{B}$ denotes a hyperparameter that determines when clustering begins. The best performance is highlighted in bold, while the worst performance is underlined. Additionally, the range of fluctuations has been indicated with $R$.}
\label{table:bgn_sensitivity}
\vspace{-1em}
\end{table}

\subsection{Analysis of Sensitivity}
\label{subsec:sensitivity}
We verify the hyperparameters sensitivity of our \method. Firstly, as we simply utilize K-means \cite{kmeans} for assigning the dynamic labels, the number of clusters \(\bm{K}\) should be given carefully consideration. It is challenging to ascertain the total number of image categories without ground truth labels, and finding the optimal parameters is computationally expensive and time-consuming working on large-scale datasets. However, the strong robustness exhibited by proposed \methodspace will alleviate this problem, as shown in \cref{table:k_sensitivity}. We validate performance of various \(\bm{K}\) on CIFAR100-LT dataset, the best item is indicated by bold text and underline for the worst one. Additionally, we report the accuracy range of each benchmark, spanning from 0.27\% to 1.23\%. Our method maintains competitive results even with K=200, and the high robustness of \(K\) helps us to find the optimal parameters more efficiently in the face of unknown datasets.

Furthermore, we further conduct sensitivity experiment of \(\bm{B}\), which determines when to start using the clustering results to influence the learning objective. Setting \(B\)  larger may harm the concentration on long-tailed learning process, while too small may have impact on the clustering results and drive the model to learn from non-ideal metrics, as shown in \cref{table:bgn_sensitivity}. We fix \(K\)=10 for experiments on the CIFAR10-LT dataset (split 1). From the results, we can see that the performance difference in \(B\) variation from 0 to 500 is minimal with the setup of 2000 epochs training.

\begin{table*}[!ht]
\centering
\normalsize
\begin{tabular}{c|cc|cccccc}  
\textbf{Method} & {$\bm{\tau}$} & {$\bm{w}$} & KNN@1 & KNN@10 & 1\%S LP & MS LP& LT LP  & Full LP\\
\midrule
SimCLR & \usym{2717} & \usym{2717} & 57.74 & 56.93 & 63.38 & 61.58 & 56.29 & 75.61\\
TS & \usym{2713} & \usym{2717}&   61.77&   61.16&  70.35&  70.06&   63.12&  78.61\\
\midrule
\multirow{3}{*}{\method}& \usym{2717}& \usym{2713}& 60.68&  60.32&  68.02&  67.34&  60.67&  77.95\\
 & \usym{2713} & \usym{2717} &  61.97&    61.52&	70.72&   70.69&  63.58&  78.97\\
 & \usym{2713} & \usym{2713} &  \textbf{63.08}&    \textbf{62.30}&	\textbf{71.78}&   \textbf{70.84}&  \textbf{64.50}&  \textbf{79.58} \\

\end{tabular}
\vspace{.25em}
\caption{Ablation Experiments. Performance comparison on CIFAR10-LT dataset (Split 1) with ResNet18. {$\bm{\tau}$} denotes employing a dynamic temperature strategy, and {$\bm{w}$} denotes using a re-weight operation. The best performance is highlighted in bold.}
\label{table:ablation}
\vspace{-1em}
\end{table*}

\begin{table}[ht]
\centering
\normalsize
\renewcommand{\arraystretch}{1.1}
\tabcolsep=0.25cm
\begin{tabular}{c|cccc|c} 
\toprule
\diagbox{$\bm{F}$}{$\bm{S}$} & 200 & 500 & 800 & 1000 & $R_F$ \\
\hline
1 & \textbf{71.49} & 70.39 & 71.16 & 71.13 & 1.1 \\
5 & 70.22 & 70.29 & 70.14 & 70.23 & 0.15\\
10 & 70.70 & 70.84 & 70.65 & 70.48 & 0.36\\
30 & 70.97 & 71.21 & 70.54 & 70.97 & 0.67\\
\hline
\rule{0pt}{3ex}
$R_S$ &1.27 &0.92 &1.02 &0.90&\diagbox{}{} \\
\bottomrule
\end{tabular}
\vspace{.25em}
\caption{The 1\%S LP performance of varying $\bm{F}$ and $\bm{S}$ on CIFAR10-LT dataset with ResNet18. $\bm{F}$ denotes the frequent of clustering, while $\bm{S}$ denotes epochs of warming. The best performance is highlighted in bold. $R_F$ and $R_S$  denote the range of fluctuations of accuracy with fixed $\bm{F}$ and $\bm{S}$, respectively.}
\label{table:fs_sensitivity}
\vspace{-1em}
\end{table}

Beyond that, we also conduct more sensitivity validation of clustering hyper-parameters shown as \cref{table:fs_sensitivity}, \ie, clustering frequency \(F\) and clustering warming epochs \(S\). We report the 1\%S LP performance on CIFAR10-LT split 1 with \(K\)=10 and \(B\)=50, and  bold the best accuracy. The results demonstrate that the performance remained consistent across a range of parameter combinations. 
The proposed \methodspace displays a great robustness across multiple settings, thereby reducing the additional tuning overhead typically associated with transferring to other public datasets.
 
\noindent
\subsection{Ablation Study}
\label{subsec:ablaion}
To verify the importance of each component of proposed \method, we conduct ablation experiments for proposed component, as shown in \cref{table:ablation}. Our approach has adaptive temperature awareness and quantity-aware re-weighting, both of which utilise the temporary labels obtained from clustering to correct the neglect of tail classes.

Compared to SimCLR, our method maintains a consist gain from clustering information. \methodspace improves the performances on all the benchmarks ranging from 3.97\% to 9.26\%, with an average improvement of 6.75\%. Even only utilising the re-weighting  strategy, \methodspace can also have an average of 3.90\% improvement.

In contrast to the  TS using temperature cosine variation strategy, \methodspace still improves accuracy by about 0.4\% on each benchmark when only using temperature adaptation. These results corroborate the effectiveness of our proposed clustering enhancement approach.

\begin{figure}[h]
    \centering
    \begin{subfigure}[b]{0.23\textwidth}
        \centering
        \includegraphics[width=\textwidth]{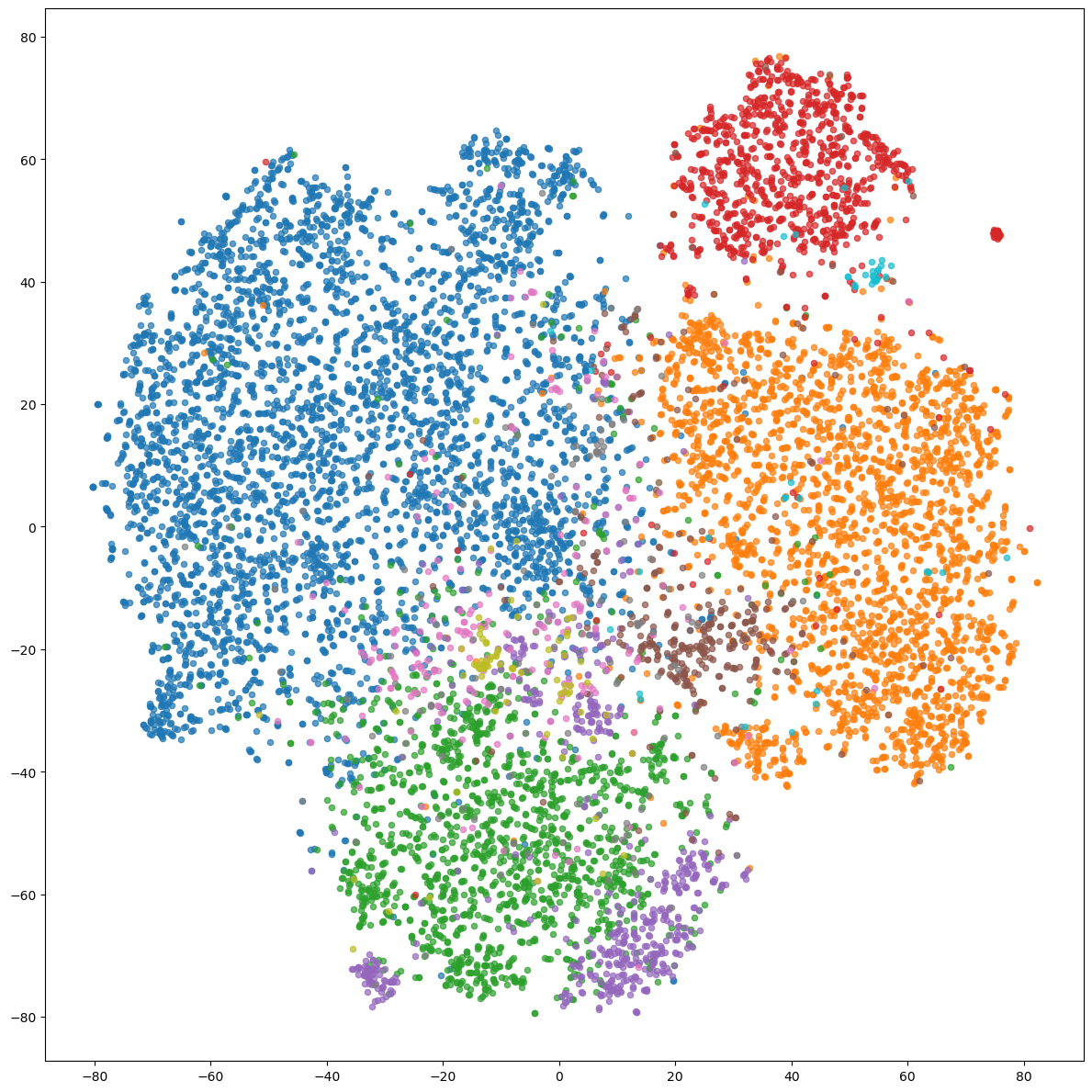}
        \caption*{\textbf{SDCLR} \cite{SDCLR}}
    \end{subfigure}
    \hfill
    \begin{subfigure}[b]{0.23\textwidth}
        \centering
        \includegraphics[width=\textwidth]{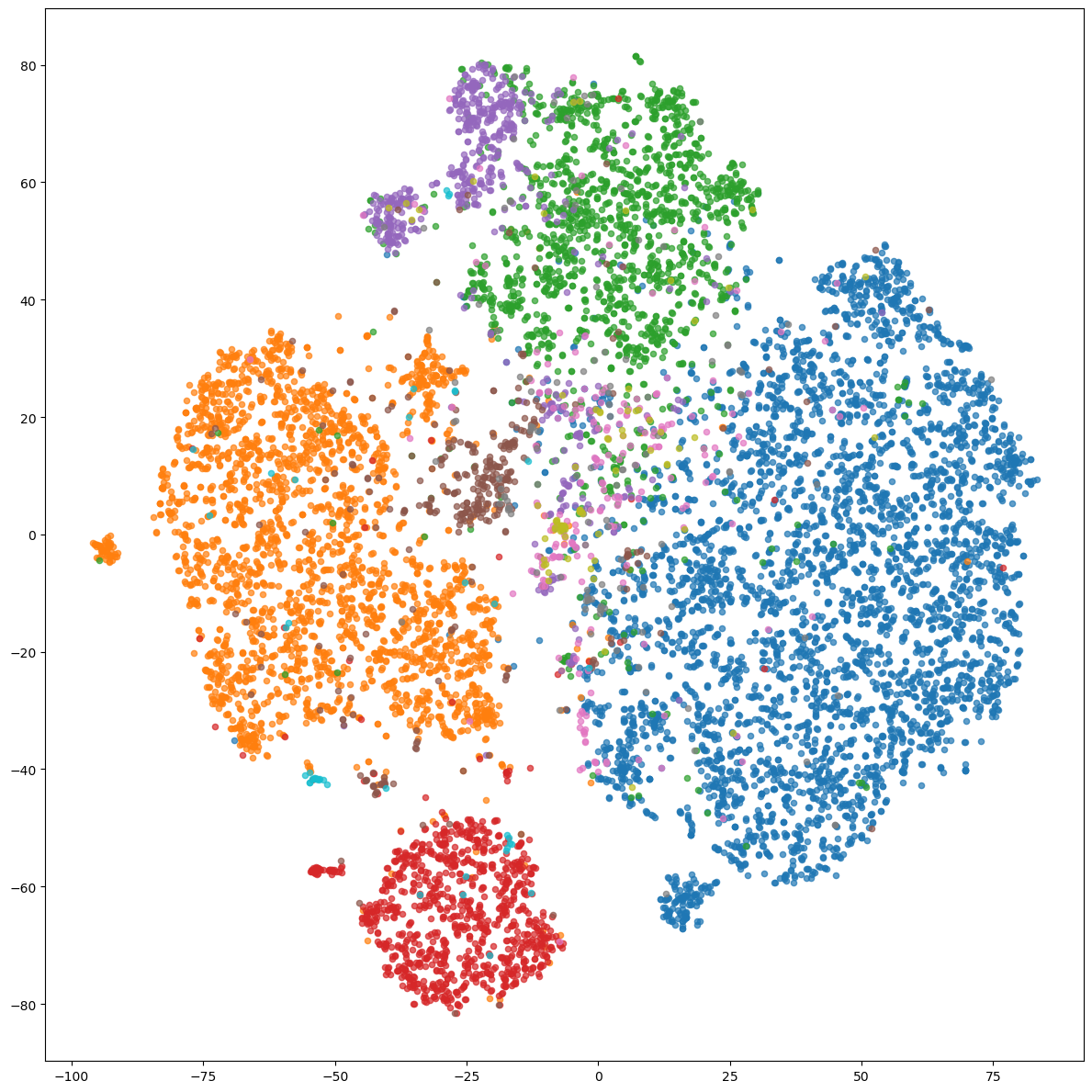}
        \caption*{\textbf{\methodspace (Ours)}}
    \end{subfigure}
    \caption{T-SNE Visualization. Different colors indicate different classes.}
    \label{fig:tsne}
\end{figure}

\noindent
\subsection{T-SNE Visualization}
\label{subsec:tsne}
T-SNE provides a qualitative perspective into the learned discriminative representation. As shown in \cref{fig:tsne}, we select samples from all 10 classes of CIFAR10-LT to generate plots. Compared to the SDCLR, our \methodspace exhibits tighter clusters and more clearly distinguishing boundaries. The results demonstrate that our method is effective in producing high-quality auxiliary supervision when facing long-tailed distributions.

\section{Conclusion}
In this paper, we proposed a novel method for self-supervised long-tailed recognition called \method.
We introduce pseudo-label assignment to drive the dynamic temperature strategy and re-weighting to counteract the long-tailed distribution, training different samples with optimal training modes to obtain robust representations.
Our approach demonstrates that the experience of supervised long-tail learning can be migrated to self-supervised long-tail learning with low overhead, in contrast to previous work that proposed non-generic solutions due to the absence of labels.
We conduct various benchmarks on multiple datasets to demonstrate the efficiency and robustness of our proposed method. 
T-SNE \cite{tsne} visualization analysis also provides a perspective showing better separability of our local representations.
{
    \small
    \bibliographystyle{ieeenat_fullname}
    \bibliography{main}
}

\clearpage
\setcounter{page}{1}
\maketitlesupplementary

\section{Other Experimental Results}

\subsection{Fine-grained Results}
\label{sec:detail}

We also summarize a detailed results of CIFAR10-LT and ImageNet100-LT to verify the performance of fine-grained classes in \cref{table:detail_cifar10} and \cref{table:detail_imagenet100}.

We provide accuracy for different categories.The situation of split is referenced in \cref{subsec:setup}.

\subsection{Results on Balanced Datasets}
\label{sec:balance}

We also performed the same experiments on the balanced dataset, as shown in \cref{table:balance}.

The results show that our method can achieve a significant lead on most benchmarks, and a similar performance to the previous work on Full LP. This indicates that our method is effective in identifying potential subclasses within a balanced dataset, thereby enhancing the performance.

\begin{table}[ht]
\centering
\small
\tabcolsep=0.12cm
\begin{tabular}{c|cccccc}  
\textbf{Method} & K@1 & K@10 &MS LP& 1\%S LP & LT LP  & Full LP\\
\midrule
SimCLR & 53.14 & 47.15 & 67.86 & 74.81 & 61.85&80.69\\
SDCLR & 64.01&60.50&76.38&75.81&71.42&\textbf{84.75}\\
TS & 65.33&62.95&76.99&78.56&69.70&83.32\\
\method & \textbf{69.61}&\textbf{67.44}&\textbf{78.27}&\textbf{79.02}&\textbf{71.77}&84.15\\
\end{tabular}
\vspace{.25em}
\caption{Performance comparison on balanced CIFAR10 dataset with ResNet18. }
\label{table:balance}
\vspace{-1em}
\end{table}

\subsection{Results via Other Clustering Methods}
\label{sec:cluster}
Alternative clustering methods are employed to implement our \method, result is summarized in \cref{table:cluster}.
Experimental results indicate that K-means and agglomerative clustering exhibit comparable performance.
More clustering methods such as discriminative clustering will be explored in future work.
\begin{table}[ht]
\centering
\small
\tabcolsep=0.12cm
\begin{tabular}{c|cccccc}  
\textbf{Method} & K@1 & K@10 &MS LP& 1\%S LP & LT LP  & Full LP\\
\midrule
DBSCAN & 60.96&60.77&69.74&68.97&61.83&78.67\\
A. C. & 62.22&61.49&71.57&\textbf{71.60}&64.39&79.46\\
K-means & \textbf{63.08} & \textbf{62.30} & \textbf{71.78} & 70.84 & \textbf{64.50} &\textbf{79.58}\\
\end{tabular}
\vspace{.25em}
\caption{Effect of different clustering methods on CIFAR10-LT dataset (Split 1) with ResNet18. A. C. denotes Agglomerative Clustering.}
\label{table:cluster}
\vspace{1em}
\end{table}

\section{Datasets Source}
\label{sec:datasets}
We used three datasets: CIFAR10-LT, CIFAR100-LT, and ImageNet100-LT. They were all obtained by long-tail subset sampling from the original datasets, namely CIFAR10, CIFAR100, and ImageNet. Sources of original datasets can be found in \cref{table:Datasource}.
These datasets are publicly available and widely used. All data are in full compliance with all applicable ethical standards.

\section{Computing Infrastructure}
Our codes are based on Pytorch 2.4.0. Our models and reproductions of previous work are trained on GeForce RTX 3090 and GeForce RTX 4090.

\begin{table*}[ht]
\centering
\normalsize
\begin{tabular}{c|cccc|cccc}  
\toprule
& \multicolumn{4}{c|}{KNN@1} &  \multicolumn{4}{c}{KNN@10} \\
Method & Head & Mid & Tail & All & Head & Mid & Tail & All\\
\midrule
SimCLR   & 87.68\small$\pm$4.28 & 62.46\small$\pm$12.35 & 29.01\small$\pm$15.3 & 62.51 & 90.68\small$\pm$4.09 & 62.14\small$\pm$12.77 & 21.91\small$\pm$18.21 & 61.49 \\
SDCLR & 86.92\small$\pm$4.25&	62.36\small$\pm$12.15&	30.38\small$\pm$13.37&	62.59&
90.71\small$\pm$4.24&	61.50\small$\pm$12.49&	21.78\small$\pm$16.83&	61.27\\
TS & 88.39\small$\pm$3.99&	61.91\small$\pm$13.02&	30.59\small$\pm$14.46&	63.11&
91.46\small$\pm$3.84&	61.26\small$\pm$12.60&	23.26\small$\pm$17.07&	61.94\\
\method & \textbf{88.63}\small$\pm$3.85&	\textbf{64.15}\small$\pm$11.74&	\textbf{31.58}\small$\pm$14.09&	\textbf{64.17}&
\textbf{91.86}\small$\pm$3.66&	\textbf{63.11}\small$\pm$12.03&	\textbf{23.84}\small$\pm$17.50&	\textbf{62.83}\\
\midrule
\midrule
& \multicolumn{4}{c|}{MS LP}&  \multicolumn{4}{c}{1\%S LP}  \\
Method & Head & Mid & Tail & All & Head & Mid & Tail & All \\
\midrule
SimCLR   & 79.48\small$\pm$5.30 & 70.49\small$\pm$11.73 & 66.32\small$\pm$12.48 & 72.83& 79.63\small$\pm$4.12 & 71.31\small$\pm$14.46 & 63.09\small$\pm$11.5 & 72.17 \\
SDCLR & 77.75\small$\pm$6.21&	73.10\small$\pm$10.82&	67.57\small$\pm$13.71&	73.30&
77.11\small$\pm$5.97	&71.91\small$\pm$13.98&	\textbf{67.71}\small$\pm$11.34&	72.73\\
TS &	79.15\small$\pm$9.23&	\textbf{73.65}\small$\pm$12.08&	65.97\small$\pm$15.20&	73.54&
79.38\small$\pm$5.72	&72.51\small$\pm$13.32&	65.93\small$\pm$11.21&	73.28\\
\method &	\textbf{79.61}\small$\pm$7.34&	73.27\small$\pm$11.52&	\textbf{67.79}\small$\pm$10.43&	\textbf{74.16}&
\textbf{80.30}\small$\pm$5.23&	\textbf{73.53}\small$\pm$12.57&	66.96\small$\pm$12.41&	\textbf{74.27}\\
\midrule
\midrule
&  \multicolumn{4}{c|}{LT LP} &  \multicolumn{4}{c}{Full LP}  \\
Method & Head & Mid & Tail & All & Head & Mid & Tail & All \\
\midrule
SimCLR& 93.04\small$\pm$2.95 & 66.89\small$\pm$11.95 & 28.34\small$\pm$14.20 & 65.78  & 87.60\small$\pm$4.67 & 83.35\small$\pm$6.92 & 79.01\small$\pm$7.77 & 83.75  \\
SDCLR & 93.68\small$\pm$3.24&	69.53\small$\pm$9.80&	29.15\small$\pm$13.37&	67.07&
\textbf{89.50}\small$\pm$3.42	&\textbf{85.20}\small$\pm$6.00&	80.77\small$\pm$7.01&	\textbf{85.59}\\
TS&	93.70\small$\pm$2.73&	68.98\small$\pm$11.82&	28.75\small$\pm$12.30&	66.80&
88.00\small$\pm$4.54&	84.41\small$\pm$6.65&	80.79\small$\pm$6.31	&84.76\\
\method&	\textbf{94.14}\small$\pm$2.52&	\textbf{70.77}\small$\pm$9.30&	\textbf{30.86}\small$\pm$12.94&	\textbf{68.15}&
89.03\small$\pm$4.23&	85.03\small$\pm$6.69&	\textbf{81.12}\small$\pm$7.05&	85.46\\
\bottomrule
\end{tabular}
\caption{Detailed results on CIFAR10-LT with ResNet50.}
\label{table:detail_cifar10}
\end{table*}

\begin{table*}[ht]
\centering
\normalsize
\begin{tabular}{c|cccc|cccc|cccc}  
\toprule
& \multicolumn{4}{c|}{KNN@1} &  \multicolumn{4}{c|}{KNN@10} &  \multicolumn{4}{c}{1\%S LP}  \\
Method & Head & Mid & Tail & All & Head & Mid & Tail & All & Head & Mid & Tail & All\\
\midrule
SimCLR   & 55.13 & 30.00 & 10.71 & 37.10 & 58.51 & 29.70 & 8.71 & 38.00 & 51.79 & 36.77 & 30.29 & 42.16 \\
SDCLR & 55.74&	27.67&	10.15&	36.06&	60.68&	26.00&	5.23&	37.36&	48.31&	39.17&	36.46&	42.38 \\
TS & 57.23&	30.26&	13.14&	38.36&	60.41&	29.53&	10.14&	38.86&	60.41&	40.38&	33.57	&45.18\\
\method-800 & 59.26	& 31.84 & 10.92 &	\textbf{39.54}&	64.11&	30.24&	8.31&	\textbf{40.26}&	54.16&	40.78&	37.85&	\textbf{45.48}\\
\method-1700 & \textit{61.53}	&\textit{32.90}&	\textit{12.00}&	\textit{41.06}&	\textit{65.63}&	\textit{32.69}&	\textit{8.46}&	\textit{42.06}&	\textit{56.16}&	\textit{43.43}&	\textit{38.77}&	\textit{47.66}\\
\midrule
&  \multicolumn{6}{c|}{LT LP} &  \multicolumn{6}{c}{Full LP}  \\
Method & \multicolumn{6}{c|}{\hspace{-0.15cm}Head\hspace{0.9cm}Mid\hspace{0.85cm}Tail\hspace{1cm}All}&\multicolumn{6}{c}{\hspace{-0.2cm}Head\hspace{0.9cm}Mid\hspace{0.95cm}Tail\hspace{1.05cm}All} \\
\midrule
SimCLR & \multicolumn{6}{c|}{67.59\hspace{0.75cm}36.47\hspace{0.75cm}9.43\hspace{0.75cm}44.82}& \multicolumn{6}{c}{69.54\hspace{0.75cm}63.71\hspace{0.75cm}59.69\hspace{0.75cm}65.46}\\
SDCLR & \multicolumn{6}{c|}{71.74\hspace{0.75cm}36.86\hspace{0.75cm}8.00\hspace{0.75cm}46.40}& \multicolumn{6}{c}{70.10\hspace{0.75cm}65.04\hspace{0.75cm}60.92\hspace{0.75cm}66.48}\\
TS & \multicolumn{6}{c|}{70.67\hspace{0.75cm}38.85\hspace{0.65cm}10.29\hspace{0.68cm}47.26}& \multicolumn{6}{c}{72.53\hspace{0.75cm}56.27\hspace{0.75cm}63.69\hspace{0.75cm}67.82}\\
\method-800 & \multicolumn{6}{c|}{71.16\hspace{0.75cm}39.88\hspace{0.75cm}9.54\hspace{0.75cm}\textbf{47.82}}& \multicolumn{6}{c}{72.74\hspace{0.75cm}66.41\hspace{0.75cm}62.46\hspace{0.75cm}\textbf{68.30}}\\
\method-1700 & \multicolumn{6}{c|}{\textit{71.47}\hspace{0.75cm}\textit{40.41}\hspace{0.75cm}\textit{8.92}\hspace{0.75cm}\textit{48.12}}& \multicolumn{6}{c}{\textit{73.79}\hspace{0.75cm}\textit{66.53}\hspace{0.75cm}\textit{65.08}\hspace{0.75cm}\textit{69.10}}\\
\bottomrule
\end{tabular}
\caption{Detailed results on ImageNet100-LT with ResNet50.\method-800 denotes the result of training our method for 800 epochs of fair comparison. \method-1700 train our method for 1700 epochs to verify scaling performance marked in italics.}
\label{table:detail_imagenet100}
\end{table*}

\begin{table*}[ht]
\centering
\small
\tabcolsep=0.2cm
\begin{tabular}{c|c}
\textbf{Dataset} & \textbf{Source Link}\\
\midrule
CIFAR10 & https://www.cs.toronto.edu/ kriz/cifar-10-python.tar.gz \\
CIFAR100 & https://www.cs.toronto.edu/ kriz/cifar-100-python.tar.gz \\
ImageNet & http://image-net.org/download \\
\end{tabular}
\vspace{.25em}
\caption{The source of the dataset we use.}
\label{table:Datasource}
\vspace{-1em}
\end{table*}

\end{document}